\documentclass[conference]{IEEEtran}
\IEEEoverridecommandlockouts
\usepackage{cite}
\usepackage{amsmath,amssymb,amsfonts}
\usepackage{algorithmic}
\usepackage{graphicx}
\usepackage{textcomp}
\usepackage{multirow}
\usepackage[table]{xcolor}
\usepackage[hidelinks]{hyperref}
\usepackage{subcaption}
\def\BibTeX{{\rm B\kern-.05em{\sc i\kern-.025em b}\kern-.08em
    T\kern-.1667em\lower.7ex\hbox{E}\kern-.125emX}}
\begin{document}

\title{Toward Efficient Weakly Supervised Semantic Segmentation Using Only Low-Magnification Histopathological Images}
\author{
\IEEEauthorblockN{
Dung Minh Do\textsuperscript{1,2,*}, 
Nhat-Thanh Huynh\textsuperscript{1,2,*}, 
Duc Minh Huynh\textsuperscript{1,2}, 
Doanh C. Bui\textsuperscript{1,2}, and 
Khang Nguyen\textsuperscript{1,2,\ensuremath{\dagger}}
}
\IEEEauthorblockA{
\textsuperscript{1}\textit{University of Information Technology, Ho Chi Minh City, Vietnam}
}
\IEEEauthorblockA{
\textsuperscript{2}\textit{Vietnam National University Ho Chi Minh City, Ho Chi Minh City, Vietnam}
}
\IEEEauthorblockA{
\{23520326, 23521440\}@gm.uit.edu.vn, 
\{duchm, doanhbc, khangnttm\}@uit.edu.vn
}
\IEEEauthorblockA{
\textsuperscript{*}These authors contributed equally to this work.
}
\IEEEauthorblockA{
\textsuperscript{\ensuremath{\dagger}}Corresponding author: khangnttm@uit.edu.vn.
}
}


\maketitle

\begin{abstract}
Whole-slide images (WSIs) provide rich tissue-level and cellular-level information, but storing and transmitting high-magnification pathology data is resource-intensive. Moreover, annotating WSIs at the pixel level is labor-intensive and time-consuming. Therefore, it is important to investigate whether low-magnification pathology images with limited annotations (i.e., image-level instead of pixel-level labels) can achieve performance comparable to high-magnification images.
This paper presents a systematic benchmark study on weakly supervised histopathological image segmentation under different low-resolution storage settings. Starting from high-resolution image patches, we simulate lower-magnification inputs and reconstruct them to the original size using interpolation and deep learning-based reconstruction methods before applying the weakly-supervised segmentation pipeline. This framework enables a quantitative evaluation of how weakly supervised methods respond to different levels of resolution degradation.
Experimental results show that reconstruction quality metrics alone are insufficient to predict downstream segmentation performance. In particular, the study identifies a critical degradation point where the localization of small-scale structures declines significantly. These findings provide practical guidance for designing efficient digital pathology storage systems while maintaining reliable automated analysis. Code is available at \url{https://github.com/Dung-Dx/LowMagWSS}
\end{abstract}

\section{Introduction}
The rapid adoption of digital pathology has led to a substantial increase in whole-slide image (WSI) data. High-magnification pathology images preserve detailed tissue morphology, cellular structures, and boundary information that are important for computational pathology tasks such as tissue classification and semantic segmentation \cite{amgad2019structured}. However, storing and transferring high-resolution pathology images requires considerable storage capacity and bandwidth. This creates a practical need for storage-efficient computational pathology, where images may be stored at lower magnification while still supporting reliable downstream analysis.

Weakly supervised semantic segmentation (WSS) is an attractive setting for computational pathology because it reduces the need for dense pixel-level annotations. Instead of requiring detailed masks, WSS methods can learn from weaker labels such as image-level or patch-level classification labels \cite{han2022multi,fang2024hisynseg}. However, WSS often depends on indirect localization cues, such as class activation maps or pseudo masks, which may be sensitive to image degradation \cite{zhou2016learning,selvaraju2017grad}. When pathology images are stored at lower resolution, fine tissue boundaries and local textures may be lost, potentially reducing the quality of weak supervision and the final segmentation results.

Super-resolution (SR) provides a possible way to reconstruct high-resolution pathology images from low-resolution inputs. Simple interpolation methods are computationally cheap but cannot recover lost high-frequency details. Learned pathology-oriented SR methods may generate sharper images, but visual improvement does not necessarily imply better downstream segmentation performance. This concern is particularly important in pathology, where inaccurate or hallucinated biological details may affect model predictions \cite{ma2020pathsrgan,chen2024star}.

In this paper, we evaluate whether SR reconstruction can support storage-efficient weakly supervised histopathological segmentation. We simulate low-resolution storage by downsampling high-magnification pathology images, reconstruct them using SR methods, and evaluate the reconstructed images with a fixed WSS pipeline. Our study focuses on downstream segmentation performance rather than reconstruction quality alone. \textbf{To the best of our knowledge, this is the first study that focuses on both low-magnification pathology images for reducing storage requirements and weakly supervised segmentation for reducing annotation effort.}

The main contributions of this work are as follows:
\begin{itemize}
    \item We introduce an experimental protocol for evaluating storage-efficient weakly supervised segmentation in histopathology.
    \item We compare simple interpolation with a pathology-oriented SR method under the same WSS pipeline.
    \item We analyze degradation-dependent and tissue-dependent effects of low-resolution storage, including failure cases for fine-grained tissue localization.
\end{itemize}
\section{Related Work}

\subsection{Weakly Supervised Segmentation in Histopathology}

Semantic segmentation is an important task in computational pathology because it enables pixel-level or region-level analysis of tissue components. Fully supervised segmentation models, such as U-Net and DeepLabV3+, have been widely used in biomedical and semantic segmentation tasks.
However, fully supervised training requires dense pixel-level annotations, which are expensive and time-consuming to obtain in histopathology \cite{han2022multi}.

Weakly supervised semantic segmentation reduces this annotation burden by using cheaper supervision, such as image-level or patch-level labels. Many WSS methods rely on class activation mapping (CAM), which extracts localization cues from classification networks \cite{zhou2016learning}. Grad-CAM further generalizes this idea by using gradient information to produce visual localization maps \cite{selvaraju2017grad}. Although CAM-based localization is effective, it often highlights only the most discriminative regions and may fail to cover complete tissue areas.

Several methods have been proposed to improve weakly supervised segmentation. SEAM improves CAM-based segmentation through self-supervised equivariant attention \cite{wang2020self}, while SC-CAM explores sub-category information to reduce incomplete activation \cite{chang2020weakly}. In histopathology, WSSS-Tissue uses patch-level classification labels and multi-layer pseudo-supervision for tissue semantic segmentation \cite{han2022multi}. More recently, HisynSeg uses image-mixing synthesis and consistency regularization to improve weakly supervised histopathological segmentation \cite{fang2024hisynseg}. In this work, we study how low-resolution storage and SR reconstruction affect an existing WSS pipeline.

\subsection{Super-Resolution for Pathology Images}

Image super-resolution reconstructs a high-resolution image from a low-resolution input. In computational pathology, SR is attractive because high-resolution imaging leads to large storage and transmission costs. Traditional interpolation methods are simple and deterministic, but they cannot recover fine image details lost during downsampling. Deep learning-based SR methods can improve visual sharpness, but the reconstructed textures may not always be biologically faithful.

Pathology-oriented SR has been explored in several studies. PathSRGAN applies generative adversarial learning to cytopathological image SR \cite{ma2020pathsrgan}. STAR-RL formulates pathology image SR as a spatial-temporal hierarchical reinforcement learning problem and emphasizes interpretable reconstruction to reduce the risk of untruthful biological details \cite{chen2024star}. Other studies have also investigated the relationship between SR and segmentation in breast cancer histopathology images \cite{juhong2022super}.

Most SR studies focus on reconstruction metrics such as PSNR, SSIM, FSIM, or GMSD\cite{ma2020pathsrgan}\cite{chen2024star}. However, these metrics do not necessarily indicate whether reconstructed images preserve information needed by downstream segmentation models. Therefore, our work evaluates SR from a task-driven perspective by measuring weakly supervised segmentation performance after reconstruction.

\section{Experimental Method}

\subsection{Overview}

\begin{figure*}[t]
    \centering
    \includegraphics[width=\textwidth]{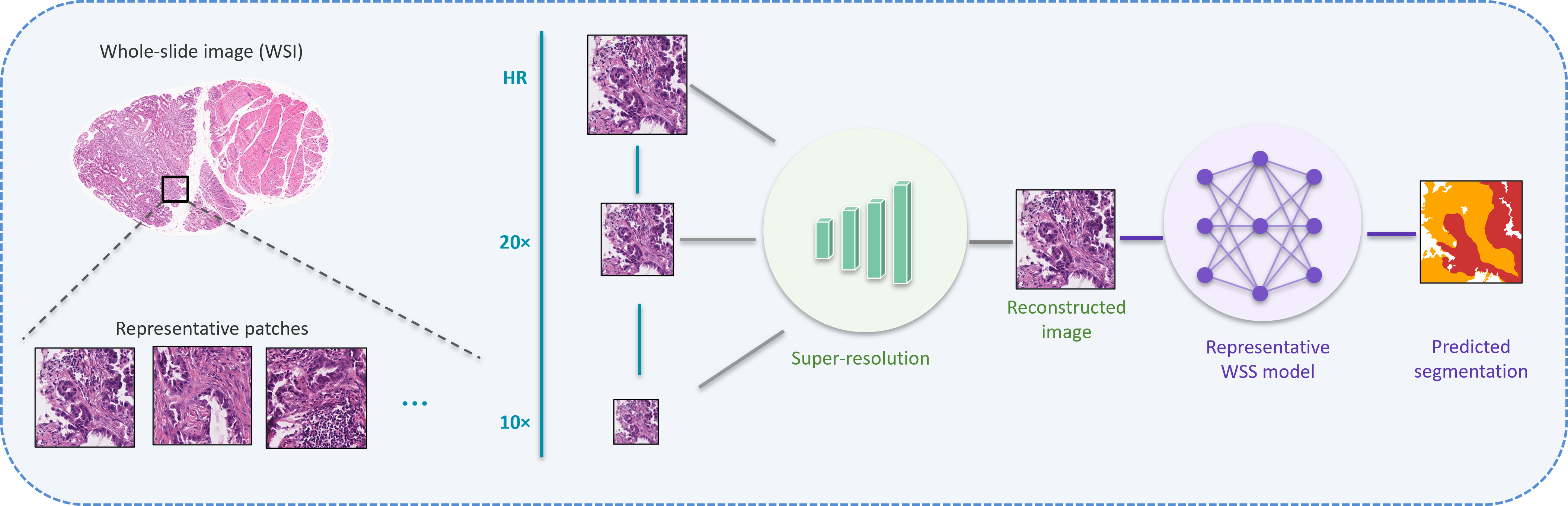}
    \caption{Overview of the proposed experimental pipeline. Original high-resolution pathology patches are downsampled to simulate low-resolution storage at 20$\times$ and 10$\times$. The degraded images are reconstructed by interpolation or super-resolution methods and then used for weakly supervised tissue segmentation.}
    \label{fig:workflow}
\end{figure*}

This study evaluates whether low-resolution storage combined with image reconstruction can preserve weakly supervised histopathological segmentation performance. The overall workflow is shown in Fig.~\ref{fig:workflow}. Given original 40$\times$ pathology patches, we simulate two storage-efficient settings by downsampling them to 20$\times$ and 10$\times$. The 20$\times$ setting reduces the spatial resolution by a factor of two along each axis, while the 10$\times$ setting reduces it by a factor of four. Therefore, the approximate pixel counts are reduced to one-fourth and one-sixteenth of the original 40$\times$ images, respectively.

The low-resolution images are reconstructed to 40$\times$ using either bilinear interpolation or STAR-RL~\cite{chen2024star}. The reconstructed images are then used as inputs to WSSS-Tissue~\cite{han2022multi}. Across all experiments, the WSS model configuration is kept unchanged; only the image resolution and reconstruction method are varied.

For each dataset, we evaluate the original 40$\times$ images and four reconstructed settings, where 20$\times$ and 10$\times$ images are restored to 40$\times$ using either bilinear interpolation or STAR-RL.

\subsection{Datasets}

Experiments are conducted on LUAD-HistoSeg and BCSS-WSSS~\cite{han2022multi}. The original 40$\times$ patches in both datasets have a spatial size of $224 \times 224$. Thus, the simulated 20$\times$ and 10$\times$ images have spatial sizes of $112 \times 112$ and $56 \times 56$, respectively. After reconstruction, all images are restored to $224 \times 224$ before segmentation.

\noindent \textbf{LUAD-HistoSeg} is a lung adenocarcinoma dataset for weakly supervised tissue semantic segmentation. The training set contains patch-level multi-label annotations, while the validation and test sets contain pixel-level masks for evaluation. Four tissue categories are used: tumor epithelial (TE), tumor-associated stroma (TAS), necrosis (NEC), and lymphocyte (LYM).

\noindent \textbf{BCSS-WSSS} is derived from the Breast Cancer Semantic Segmentation dataset~\cite{amgad2019structured}. It converts dense segmentation labels into patch-level multi-label annotations for weakly supervised training, while retaining pixel-level masks for validation and testing. Four foreground categories are evaluated: tumor (TUM), stroma (STR), lymphocytic infiltrate (LYM), and necrosis (NEC).

\subsection{Image Reconstruction Methods}

We consider two super-resolution approaches: one is the traditional bilinear interpolation method, while the other is the pathology-oriented deep learning-based super-resolution model STAR-RL \cite{chen2024star}:

\noindent\textbf{Bilinear Interpolation.} Bilinear interpolation is used as a deterministic reconstruction baseline. It upsamples each low-resolution image directly to the original 40$\times$ size. This method is simple and does not introduce learned image-specific textures, but it cannot recover high-frequency tissue structures removed during downsampling.

\noindent\textbf{STAR-RL.} STAR-RL is a pathology-oriented super-resolution method based on spatial-temporal hierarchical reinforcement learning~\cite{chen2024star}. It formulates image restoration as a Markov decision process and performs reconstruction through sequential interpretable operations. The method contains three main modules: a spatial manager for selecting degraded patches, a temporal manager for early stopping, and a patch worker for applying pixel-wise restoration actions. The action space includes operations such as edge enhancement, contrast enhancement, smoothing, sharpening, addition, subtraction, and no-operation filters.

In this experiment, STAR-RL is trained on HistoSR, which contains paired pathology images with a spatial size of $192 \times 192$~\cite{chen2024star}. The trained model is then applied to LUAD-HistoSeg and BCSS-WSSS. Because the target datasets use $224 \times 224$ patches, we use a size-matching procedure during STAR-RL inference. Each low-resolution image is first resized to $192 \times 192$ using bicubic interpolation, processed by STAR-RL, and then resized back to $224 \times 224$ using bicubic interpolation:
\[
I^{LR} \xrightarrow{\text{bicubic}} I^{LR}_{192}
\xrightarrow{\text{STAR-RL}} \hat{I}^{SR}_{192}
\xrightarrow{\text{bicubic}} \hat{I}^{SR}_{224}.
\]
Here, $I^{LR}$ denotes the simulated low-resolution image, $I^{LR}_{192}$ is the resized STAR-RL input, $\hat{I}^{SR}_{192}$ is the STAR-RL output, and $\hat{I}^{SR}_{224}$ is the final reconstructed image used for segmentation.

\subsection{Weakly Supervised Segmentation}

We use WSSS-Tissue as the weakly supervised tissue segmentation framework~\cite{han2022multi}. It learns semantic segmentation from patch-level classification labels through a two-phase pipeline. First, a multi-label classification network predicts the presence or absence of tissue categories and generates pseudo masks from class activation cues. Second, a segmentation network is trained using the generated pseudo masks.

WSSS-Tissue uses Progressive Dropout Attention to reduce the tendency of CAM-based localization to focus only on the most discriminative regions. It also uses Multi-Layer Pseudo-Supervision, where pseudo masks from multiple feature layers are used to narrow the gap between patch-level labels and pixel-level segmentation masks. A classification gate is further applied to suppress false-positive predictions for absent tissue categories.

For each input setting, the segmentation pipeline is trained and evaluated independently. The architecture, losses, hyperparameters, dataset splits, and training schedule are fixed across settings. Thus, the comparison isolates the impact of low-resolution storage and reconstruction on weakly supervised segmentation.

\subsection{Evaluation Metrics}

We evaluate segmentation performance using class-wise \textit{Intersection over Union} (IoU), mean IoU (mIoU), and frequency-weighted IoU (FwIoU). mIoU averages IoU over all tissue classes, while FwIoU weights each class by its pixel frequency. We also visualize segmentation maps to compare the effects of different reconstruction methods on tissue boundaries and small structures.

\section{Results}
\begin{table*}[t]
\caption{Weakly supervised segmentation performance under different resolution degradation and reconstruction settings. Numbers in parentheses indicate the performance drop compared with the 40$\times$ setting.}
\label{tab:results}
\begin{center}
\resizebox{1\textwidth}{!}{\begin{tabular}{cccccccc}
\hline
\rowcolor[HTML]{EFEFEF} 
\multicolumn{8}{c}{\textbf{BCSS-WSSS Dataset}} \\ \hline
\textbf{Resolution} & \textbf{Method} & \textbf{TUM} & \textbf{STR} & \textbf{LYM} & \textbf{NEC} & \textbf{mIoU} & \textbf{FwIoU} \\ \hline
40$\times$ & -- & \textcolor{gray}{79.04} & \textcolor{gray}{71.74} & \textcolor{gray}{58.11} & \textcolor{gray}{58.11} & \textcolor{gray}{68.87} & \textcolor{gray}{73.31} \\ \hline
\multirow{2}{*}{20$\times$ $\rightarrow$ 40$\times$} 
& Bilinear Interpolation 
& 78.65 {\scriptsize($\downarrow$0.39)} 
& 71.75 {\scriptsize($\uparrow$0.01)} 
& 58.94 {\scriptsize($\uparrow$0.83)} 
& 66.80 {\scriptsize($\uparrow$8.69)} 
& 69.04 {\scriptsize($\uparrow$0.17)} 
& 73.24 {\scriptsize($\downarrow$0.07)} \\ \cline{2-8} 

& STAR-RL 
& 78.88 {\scriptsize($\downarrow$0.16)} 
& 71.28 {\scriptsize($\downarrow$0.46)} 
& 58.23 {\scriptsize($\uparrow$0.12)} 
& 65.99 {\scriptsize($\uparrow$7.88)} 
& 68.59 {\scriptsize($\downarrow$0.28)} 
& 73.03 {\scriptsize($\downarrow$0.28)} \\ \hline

\multirow{2}{*}{10$\times$ $\rightarrow$ 40$\times$} 
& Bilinear Interpolation 
& 77.23 {\scriptsize($\downarrow$1.81)} 
& 70.92 {\scriptsize($\downarrow$0.82)} 
& 56.90 {\scriptsize($\downarrow$1.21)} 
& 52.36 {\scriptsize($\downarrow$5.75)} 
& 64.35 {\scriptsize($\downarrow$4.52)} 
& 71.48 {\scriptsize($\downarrow$1.83)} \\ \cline{2-8} 

& STAR-RL 
& 77.13 {\scriptsize($\downarrow$1.91)} 
& 72.31 {\scriptsize($\uparrow$0.57)} 
& 57.33 {\scriptsize($\downarrow$0.78)} 
& 56.60 {\scriptsize($\downarrow$1.51)} 
& 65.84 {\scriptsize($\downarrow$3.03)} 
& 72.25 {\scriptsize($\downarrow$1.06)} \\ \hline

\multicolumn{8}{c}{} \\ [-0.5em] \hline
\rowcolor[HTML]{EFEFEF} 
\multicolumn{8}{c}{\textbf{LUAD-HistoSeg Dataset}} \\ \hline
\textbf{Resolution} & \textbf{Method} & \textbf{TE} & \textbf{NEC} & \textbf{LYM} & \textbf{TAS} & \textbf{mIoU} & \textbf{FwIoU} \\ \hline

40$\times$ & -- 
& \textcolor{gray}{77.02} 
& \textcolor{gray}{78.41} 
& \textcolor{gray}{74.59} 
& \textcolor{gray}{71.08} 
& \textcolor{gray}{75.27} 
& \textcolor{gray}{74.59} \\ \hline

\multirow{2}{*}{20$\times$ $\rightarrow$ 40$\times$} 
& Bilinear Interpolation 
& 74.04 {\scriptsize($\downarrow$2.98)} 
& 70.12 {\scriptsize($\downarrow$8.29)} 
& 66.56 {\scriptsize($\downarrow$8.03)} 
& 68.45 {\scriptsize($\downarrow$2.63)} 
& 69.79 {\scriptsize($\downarrow$5.48)} 
& 70.67 {\scriptsize($\downarrow$3.92)} \\ \cline{2-8} 

& STAR-RL 
& 76.94 {\scriptsize($\downarrow$0.08)} 
& 68.97 {\scriptsize($\downarrow$9.44)} 
& 71.65 {\scriptsize($\downarrow$2.94)} 
& 70.77 {\scriptsize($\downarrow$0.31)} 
& 72.08 {\scriptsize($\downarrow$3.19)} 
& 73.35 {\scriptsize($\downarrow$1.24)} \\ \hline

\multirow{2}{*}{10$\times$ $\rightarrow$ 40$\times$} 
& Bilinear Interpolation 
& 43.27 {\scriptsize($\downarrow$33.75)} 
& 30.43 {\scriptsize($\downarrow$47.98)} 
& 0.00 {\scriptsize($\downarrow$74.59)} 
& 47.68 {\scriptsize($\downarrow$23.40)} 
& 30.35 {\scriptsize($\downarrow$44.92)} 
& 38.16 {\scriptsize($\downarrow$36.43)} \\ \cline{2-8} 

& STAR-RL 
& 51.57 {\scriptsize($\downarrow$25.45)} 
& 38.23 {\scriptsize($\downarrow$40.18)} 
& 0.00 {\scriptsize($\downarrow$74.59)} 
& 47.75 {\scriptsize($\downarrow$23.33)} 
& 34.39 {\scriptsize($\downarrow$40.88)} 
& 42.25 {\scriptsize($\downarrow$32.34)} \\ \hline

\end{tabular}}
\end{center}
\end{table*}





\begin{figure*}[htbp]
    \centering

    \begin{subfigure}[b]{1\linewidth}
        \centering
        \includegraphics[width=\linewidth]{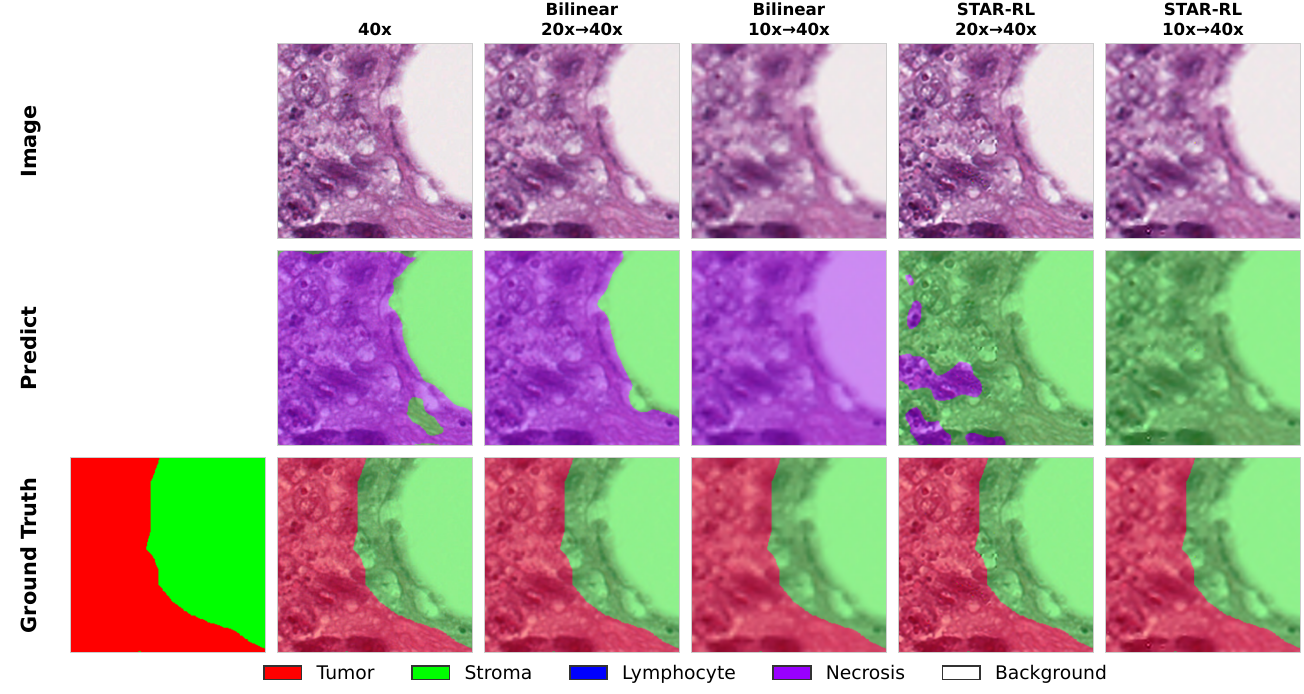}
        \caption{BCSS-WSSS}
        \label{fig:qual_bcss}
    \end{subfigure}

    \vspace{0.6em}

    \begin{subfigure}[b]{1\linewidth}
        \centering
        \includegraphics[width=\linewidth]{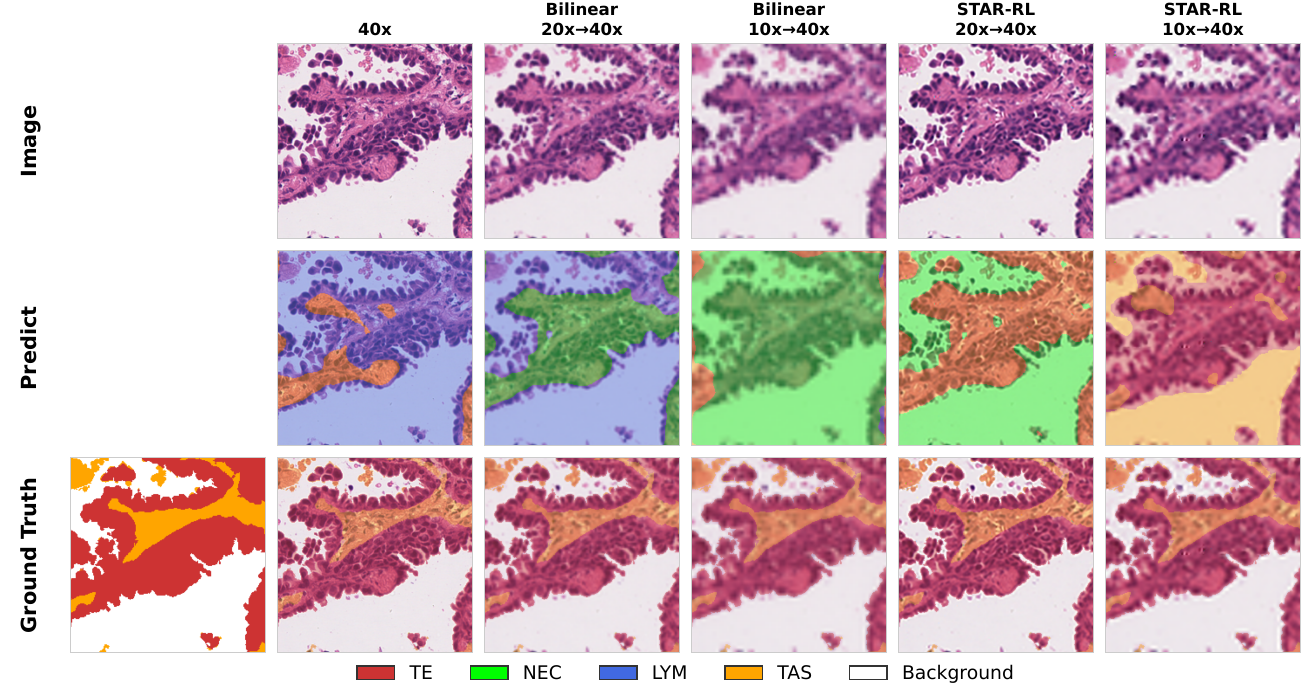}
        \caption{LUAD-HistoSeg}
        \label{fig:qual_luad}
    \end{subfigure}

    \caption{
    Qualitative weakly supervised segmentation results under different resolution degradation and reconstruction settings on BCSS-WSSS and LUAD-HistoSeg.}

    \label{fig:qual}
\end{figure*}

Table~\ref{tab:results} summarizes the segmentation performance across all five input settings on BCSS-WSSS and LUAD-HistoSeg. Fig.~\ref{fig:qual} provides qualitative visualization of the corresponding segmentation outputs under different reconstruction settings. The 40$\times$ baseline generally provides the strongest overall reference performance, although mild degradation can occasionally yield comparable or slightly higher scores for specific settings. The reconstructed settings are analyzed separately by degradation level.

\subsection{Moderate Degradation (20$\times$ $\rightarrow$ 40$\times$)}

On BCSS-WSSS, both reconstruction methods perform comparably to the 40$\times$ baseline. Bilinear interpolation achieves an mIoU of 69.04\% and FwIoU of 73.24\%, marginally surpassing the baseline of 68.87\% and 73.31\%, respectively. STAR-RL yields slightly lower overall scores (mIoU: 68.59\%, FwIoU: 73.03\%), though the difference is within a narrow margin. Notably, the necrosis class improves substantially under 20$\times$ bilinear interpolation (from 58.11\% to 66.80\%), suggesting that mild smoothing introduced by interpolation may occasionally benefit this class, although further analysis is needed to confirm this effect. These results indicate that a twofold reduction in spatial resolution can be tolerated with negligible downstream performance loss on BCSS-WSSS.

On LUAD-HistoSeg, the same moderate degradation induces a more pronounced performance gap. Bilinear interpolation reduces mIoU by 5.48 percentage points (from 75.27\% to 69.79\%) and FwIoU by 3.92 points (from 74.59\% to 70.67\%). STAR-RL substantially narrows this gap, recovering mIoU to 72.08\% and FwIoU to 73.35\%, which represents improvements of 2.29 and 2.68 points over bilinear interpolation, respectively. The most pronounced class-wise gain from STAR-RL is observed for the lymphocyte category (+5.09 IoU over bilinear at 20$\times$), suggesting that learned reconstruction better preserves fine-grained cellular structures compared to interpolation.

\subsection{Severe Degradation (10$\times$ $\rightarrow$ 40$\times$)}

At the 10$\times$ setting, which reduces the pixel count to one-sixteenth of the original, performance degrades substantially on both datasets. On BCSS-WSSS, bilinear interpolation yields an mIoU of 64.35\% and FwIoU of 71.48\%, representing drops of 4.52 and 1.83 points from the 20$\times$ bilinear setting. STAR-RL partially compensates, recovering mIoU to 65.84\% and FwIoU to 72.25\%. The largest class-wise recovery is observed for necrosis (+4.24 IoU), a tissue class characterized by irregular texture that may benefit from the edge-enhancement operations in STAR-RL's action space.

On LUAD-HistoSeg, the 10$\times$ setting causes a catastrophic performance collapse for both methods. Bilinear interpolation drops to an mIoU of 30.35\% and FwIoU of 38.16\%, while STAR-RL, despite partial recovery, still reaches only 34.39\% mIoU and 42.25\% FwIoU. The most critical finding is the complete failure of the lymphocyte class, which collapses to 0.00\% IoU under both 10$\times$ reconstruction methods. Lymphocyte regions are characterized by small, densely packed cellular structures, and the fourfold spatial reduction appears to remove the discriminative cues required by the CAM-based localization mechanism in WSSS-Tissue. This collapse is not observed on BCSS-WSSS, where the lymphocytic infiltrate class retains reasonable IoU scores (56.90\% and 57.33\% for bilinear and STAR-RL, respectively), likely due to dataset-specific differences in tissue scale and regional extent.

\subsection{Reconstruction Quality}
\begin{figure*}[htbp]
    \centering
    \includegraphics[width=\linewidth]{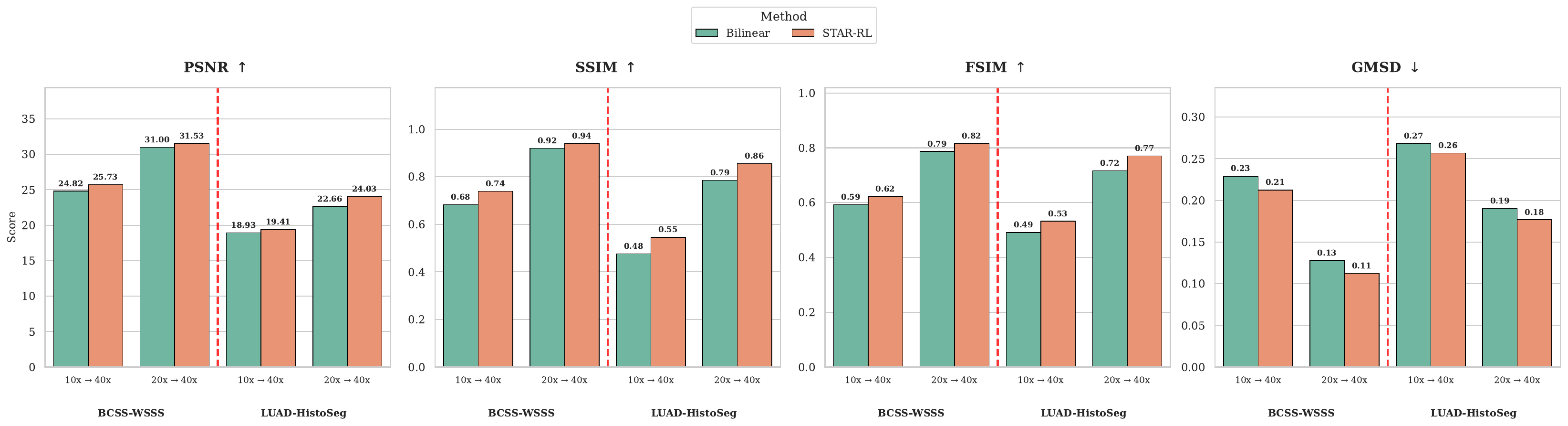}
    \caption{Reconstruction quality comparison under different degradation and reconstruction settings on BCSS-WSSS and LUAD-HistoSeg. Higher PSNR, SSIM, and FSIM indicate better reconstruction quality, while lower GMSD indicates better gradient similarity preservation.}
    \label{fig:recon_metrics}
\end{figure*}






Figure~\ref{fig:recon_metrics} presents the reconstruction quality metrics (PSNR, SSIM, FSIM, and GMSD) across the four reconstructed settings. STAR-RL consistently achieves higher PSNR, SSIM, and FSIM scores compared to bilinear interpolation, while yielding lower GMSD values (indicating better gradient similarity). Despite these reconstruction advantages, the downstream segmentation benefit of STAR-RL is dataset-dependent. On BCSS-WSSS, bilinear interpolation at 20$\times$ slightly outperforms STAR-RL in mIoU (69.04\% vs. 68.59\%), despite lower reconstruction scores. Conversely, on LUAD-HistoSeg, STAR-RL consistently outperforms bilinear interpolation across both degradation levels. These observations indicate that higher reconstruction fidelity does not linearly translate to better segmentation performance, underscoring the necessity of task-driven evaluation protocols for pathology SR methods.

\section{Discussion}

The results show that the feasibility of low-resolution storage depends on both the degradation level and the target tissue characteristics. At moderate degradation from 40$\times$ to 20$\times$, BCSS-WSSS maintains comparable performance to the original 40$\times$ setting, suggesting that large and spatially coherent tissue regions can tolerate moderate resolution loss. In contrast, LUAD-HistoSeg is more sensitive, especially for fine-grained structures such as lymphocyte regions. This indicates that the acceptable storage resolution should be determined according to the target dataset and tissue classes rather than treated as a universal setting.
At severe degradation from 40$\times$ to 10$\times$, both reconstruction methods lead to substantial performance drops. The most critical case is the lymphocyte class on LUAD-HistoSeg, where the IoU collapses to 0.00\% for both bilinear interpolation and STAR-RL. This suggests that once discriminative small-scale cues are removed during low-resolution storage, post-hoc reconstruction cannot reliably recover them for weakly supervised localization.
The comparison between reconstruction methods shows that STAR-RL is helpful in more challenging settings, particularly on LUAD-HistoSeg and under stronger degradation. However, bilinear interpolation remains competitive under mild degradation, especially on BCSS-WSSS. Therefore, the benefit of learned SR depends on the degradation level and the structural complexity of the dataset.
Another key finding is that better reconstruction quality does not always imply better segmentation performance. Although STAR-RL generally improves PSNR, SSIM, FSIM, and GMSD, these gains do not consistently translate into higher mIoU or FwIoU. This highlights the need for task-driven evaluation when applying SR methods to histopathological image analysis.


\section{Conclusions}

This paper studied whether low-resolution storage followed by reconstruction can preserve weakly supervised histopathological segmentation performance. Experiments on LUAD-HistoSeg and BCSS-WSSS show that moderate reduction to 20× can remain viable, especially for datasets with large and coherent tissue regions. In contrast, aggressive reduction to 10× causes substantial degradation and can completely remove localization cues for fine-grained classes such as lymphocytes in LUAD-HistoSeg. STAR-RL is beneficial in more challenging settings, but its advantage is not universal. The results further show that reconstruction metrics such as PSNR and SSIM are insufficient predictors of downstream segmentation quality. Therefore, low-resolution storage strategies for pathology should be selected based on downstream task performance rather than reconstruction fidelity alone.

\section*{Acknowledgment}
This research was supported by The VNUHCM-University of Information Technology's Scientific Research Support Fund.

\bibliographystyle{IEEEtran}
\bibliography{references}
\end{document}